\documentclass[10pt,twocolumn,letterpaper]{article}

\usepackage{cvpr}
\usepackage{times}
\usepackage{epsfig}
\usepackage{graphicx}
\usepackage{amsmath}
\usepackage{amssymb}
\usepackage{nicefrac}
\usepackage{mathtools}
\usepackage[acronym,shortcuts]{glossaries}
\DeclarePairedDelimiter\ceil{\lceil}{\rceil}
\DeclarePairedDelimiter\floor{\lfloor}{\rfloor}
\newcommand{\Mod}[2]{\text{mod}\left(#1,#2\right)}
\usepackage{setspace}
\usepackage[caption=false,font=footnotesize]{subfig}
\usepackage[pagebackref=true,breaklinks=true,letterpaper=true,colorlinks,bookmarks=false]{hyperref}

\usepackage{booktabs}

\cvprfinalcopy 

\def\cvprPaperID{871} 

\ifcvprfinal\fi

\makeglossaries

\newacronym{SISR}{SISR}{single image super-resolution}
\newacronym{PSNR}{PSNR}{peak signal to noise ratio}
\newacronym{MSE}{MSE}{mean squared error}
\newacronym{CNN}{CNN}{convolutional neural network}
\newacronym{ESPCN}{ESPCN}{efficient sub-pixel convolutional neural network}
\newacronym{LR}{LR}{low resolution}
\newacronym{HR}{HR}{high resolution}
\newacronym{SR}{SR}{ super-resolution}
\newacronym{HD}{HD}{high definition}
\newacronym{FPS}{FPS}{frames per second}
\newacronym{TNRD}{TNRD}{trainable nonlinear reaction diffusion}

\newcommand{\lrimage}[0]{\mathbf{I}^{LR}}
\newcommand{\hrimage}[0]{\mathbf{I}^{HR}}
\newcommand{\networkoutput}[0]{\mathbf{I}^{SR}}
\newcommand{\convolution}[2]{#1\ast#2}
\newcommand{\nonlinearity}[0]{\phi}
\newcommand{\periodicshufflingoperator}[0]{\mathcal{PS}}

\begin{document}

\title{Real-Time Single Image and Video Super-Resolution Using an Efficient Sub-Pixel Convolutional Neural Network}

\onehalfspacing
\author{Wenzhe Shi$^1$, Jose Caballero$^1$, Ferenc Husz\'{a}r$^1$, Johannes Totz$^1$, Andrew P. Aitken$^1$, \\
        Rob Bishop$^1$, Daniel Rueckert$^1$, Zehan Wang$^1$\\
        $^1$Twitter\\
        {\tt\small $^1$\{wshi,jcaballero,fhuszar,jtotz,aitken,rbishop,zehanw\}@twitter.com}
}

\maketitle

\singlespacing
\begin{abstract}
Recently, several models based on deep neural networks have achieved great success in terms of both reconstruction accuracy and computational performance for single image super-resolution. In these methods, the low resolution (LR) input image is upscaled to the high resolution (HR) space using a single filter, commonly bicubic interpolation, before reconstruction. This means that the  super-resolution (SR) operation is performed in HR space. We demonstrate that this is sub-optimal and adds computational complexity. In this paper, we present the first convolutional neural network (CNN) capable of real-time SR of 1080p videos on a single K2 GPU. To achieve this, we propose a novel CNN architecture where the feature maps are extracted in the LR space. In addition, we introduce an efficient sub-pixel convolution layer which learns an array of upscaling filters to upscale the final LR feature maps into the HR output. By doing so, we effectively replace the handcrafted bicubic filter in the SR pipeline with more complex upscaling filters specifically trained for each feature map, whilst also reducing the computational complexity of the overall SR operation. We evaluate the proposed approach using images and videos from publicly available datasets and show that it performs significantly better (+0.15dB on Images and +0.39dB on Videos) and is an order of magnitude faster than previous CNN-based methods.
\end{abstract}

\section{Introduction}

The recovery of a \ac{HR} image or video from its \ac{LR} counter part is topic of great interest in digital image processing. This task, referred to as \ac{SR}, finds direct applications in many areas such as HDTV \cite{Goto2014}, medical imaging \cite{Peled2001,shi2013cardiac}, satellite imaging \cite{Thornton2006}, face recognition \cite{Gunturk2003} and surveillance \cite{Zhang2010a}. The global \ac{SR} problem assumes \ac{LR} data to be a low-pass filtered (blurred), downsampled and noisy version of \ac{HR} data. It is a highly ill-posed problem, due to the loss of high-frequency information that occurs during the non-invertible low-pass filtering and subsampling operations. Furthermore, the SR operation is effectively a one-to-many mapping from \ac{LR} to \ac{HR} space which can have multiple solutions, of which determining the correct solution is non-trivial. A key assumption that underlies many \ac{SR} techniques is that much of the high-frequency data is redundant and thus can be accurately reconstructed from low frequency components. \ac{SR} is therefore an inference problem, and thus relies on our model of the statistics of images in question.

Many methods assume multiple images are available as \ac{LR} instances of the same scene with different perspectives, i.e. with unique prior affine transformations. These can be categorised as multi-image \ac{SR} methods \cite{Borman1998a, Farsiu2004} and exploit \emph{explicit redundancy} by constraining the ill-posed problem with additional information and attempting to invert the downsampling process. However, these methods usually require computationally complex image registration and fusion stages, the accuracy of which directly impacts the quality of the result. An alternative family of methods are \ac{SISR} techniques \cite{yang2014single}. These techniques seek to learn \emph{implicit redundancy} that is present in natural data to recover missing \ac{HR} information from a single \ac{LR} instance. This usually arises in the form of local spatial correlations for images and additional temporal correlations in videos. In this case, prior information in the form of reconstruction constraints is needed to restrict the solution space of the reconstruction.

\begin{figure*}[htbp]
\begin{center}
\includegraphics[width=1.0\linewidth]{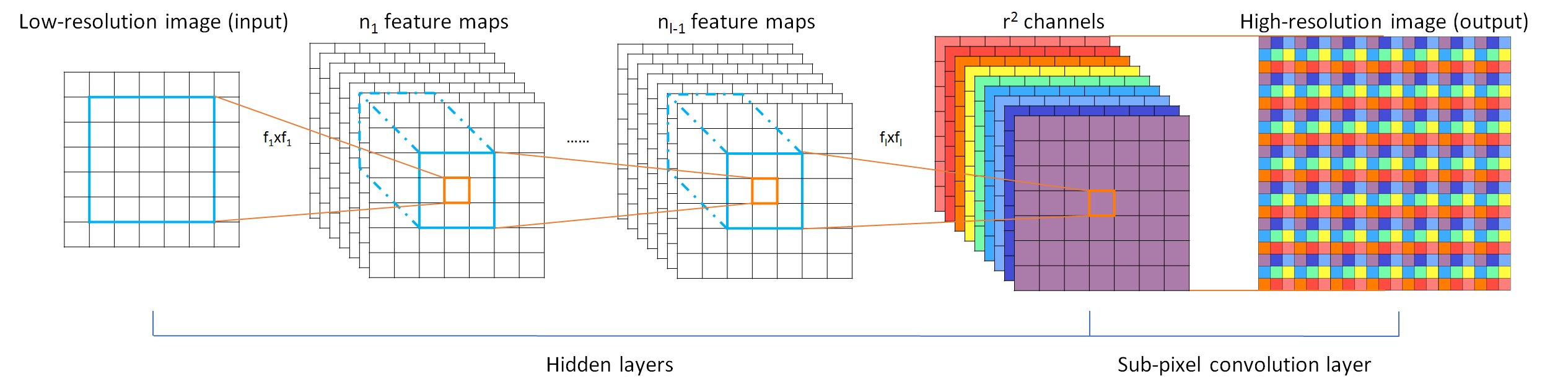}
\caption{The proposed efficient sub-pixel convolutional neural network (ESPCN), with two convolution layers for feature maps extraction, and a sub-pixel convolution layer that aggregates the feature maps from \ac{LR} space and builds the \ac{SR} image in a single step.}
\label{fig:networkstructure}
\end{center}
\end{figure*}

\subsection{Related Work}

The goal of \ac{SISR} methods is to recover a \ac{HR} image from a single \ac{LR} input image \cite{glasner2009super}. Recent popular \ac{SISR} methods can be classified into edge-based \cite{sun2011gradient}, image statistics-based \cite{efrat2013accurate,he2011single,yang2013fast,fernandez2013super} and patch-based \cite{chang2004super,wang2012semi,zhang2012multi,gao2012image,zhu2014single,timofte2014a+,dai2015jointly} methods. A detailed review of more generic \ac{SISR} methods can be found in \cite{yang2014single}. One family of approaches that has recently thrived in tackling the \ac{SISR} problem is sparsity-based techniques. Sparse coding is an effective mechanism that assumes any natural image can be sparsely represented in a transform domain. This transform domain is usually a dictionary of image atoms \cite{Mallat:2008:WTS:1525499,Elad:2010:SRR:1895005}, which can be learnt through a training process that tries to discover the correspondence between \ac{LR} and \ac{HR} patches. This dictionary is able to embed the prior knowledge necessary to constrain the ill-posed problem of super-resolving unseen data. This approach is proposed in the methods of \cite{Yang2010c, Dong2011}. A drawback of sparsity-based techniques is that introducing the sparsity constraint through a nonlinear reconstruction is generally computationally expensive.

Image representations derived via neural networks \cite{krizhevsky2012imagenet,zeiler2014visualizing,simonyan2014very} have recently also shown promise for \ac{SISR}. These methods, employ the back-propagation algorithm \cite{le1990handwritten} to train on large image databases such as ImageNet \cite{russakovsky2014imagenet} in order to learn nonlinear mappings of \ac{LR} and \ac{HR} image patches. Stacked collaborative local auto-encoders are used in \cite{cui2014deep} to  super-resolve the \ac{LR} image layer by layer. Osendorfer et al. \cite{osendorfer2014image} suggested a method for \ac{SISR} based on an extension of the predictive convolutional sparse coding framework \cite{poultney2006efficient}. A multiple layer \ac{CNN} inspired by sparse-coding methods is proposed in \cite{dong2015image}. Chen et. al. \cite{chen2015trainable} proposed to use multi-stage \ac{TNRD} as an alternative to \ac{CNN} where the weights and the nonlinearity is trainable. Wang et. al \cite{wang2015deeply} trained a cascaded sparse coding network from end to end inspired by LISTA (Learning iterative shrinkage and thresholding algorithm) \cite{gregor2010learning} to fully exploit the natural sparsity of images. The network structure is not limited to neural networks, for example, a random forest \cite{schulter2015fast} has also been successfully used for \ac{SISR}.

\subsection{Motivations and contributions}

\begin{figure}[htbp]
\begin{center}
\includegraphics[width=1.0\linewidth]{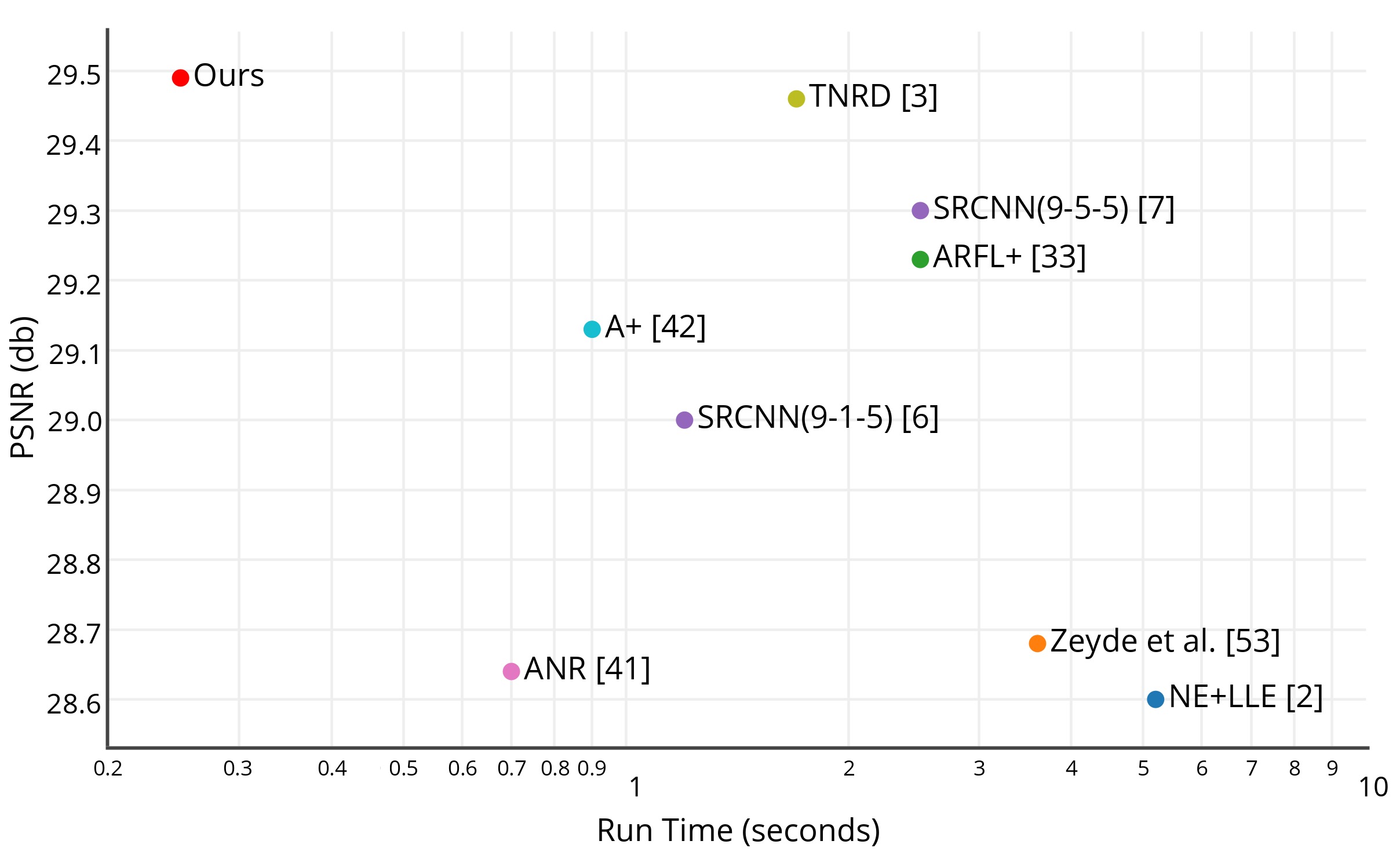}
\caption{Plot of the trade-off between accuracy and speed for different methods when performing \ac{SR} upscaling with a scale factor of 3. The results presents the mean \ac{PSNR} and run-time over the images from Set14 run on a single CPU core clocked at 2.0 GHz.}
\label{fig:accuracyvscomplexity}
\end{center}
\end{figure}

With the development of \ac{CNN}, the efficiency of the algorithms, especially their computational and memory cost, gains importance \cite{Szegedy43022}. The flexibility of deep network models to learn nonlinear relationships has been shown to attain superior reconstruction accuracy compared to previously hand-crafted models \cite{osendorfer2014image,dong2015image,wang2015deeply,schulter2015fast,chen2015trainable}. To  super-resolve a \ac{LR} image into \ac{HR} space, it is necessary to increase the resolution of the \ac{LR} image to match that of the \ac{HR} image at some point.

In Osendorfer et al. \cite{osendorfer2014image}, the image resolution is increased in the middle of the network gradually. Another popular approach is to increase the resolution before or at the first layer of the network \cite{dong2015image,wang2015deeply,chen2015trainable}. However, this approach has a number of drawbacks. Firstly, increasing the resolution of the \ac{LR} images before the image enhancement step increases the computational complexity. This is especially problematic for convolutional networks, where the processing speed directly depends on the input image resolution. Secondly, interpolation methods typically used to accomplish the task, such as bicubic interpolation \cite{dong2015image,wang2015deeply,chen2015trainable}, do not bring additional information to solve the ill-posed reconstruction problem.

Learning upscaling filters was briefly suggested in the footnote of Dong et.al. \cite{dong2014learning}. However, the importance of integrating it into the CNN as part of the SR operation was not fully recognised and the option not explored. Additionally, as noted by Dong et al. \cite{dong2014learning}, there are no efficient implementations of a convolution layer whose output size is larger than the input size and well-optimized implementations such as convnet \cite{krizhevsky2012imagenet} do not trivially allow such behaviour.

In this paper, contrary to previous works, we propose to increase the resolution from \ac{LR} to \ac{HR} only at the very end of the network and  super-resolve \ac{HR} data from \ac{LR} feature maps. This eliminates the need to perform most of the \ac{SR} operation in the far larger \ac{HR} resolution. For this purpose, we propose an efficient sub-pixel convolution layer to learn the upscaling operation for image and video super-resolution.

The advantages of these contributions are two fold:
\begin{itemize}
  \item In our network, upscaling is handled by the last layer of the network. This means each \ac{LR} image is directly fed to the network and feature extraction occurs through nonlinear convolutions in \ac{LR} space. Due to the reduced input resolution, we can effectively use a smaller filter size to integrate the same information while maintaining a given contextual area. The resolution and filter size reduction lower the computational and memory complexity substantially enough to allow super-resolution of \ac{HD} videos in real-time as shown in Sec.~\ref{subsection:runtime}.
  \item For a network with $L$ layers, we learn $n_{L-1}$ upscaling filters for the $n_{L-1}$ feature maps as opposed to one upscaling filter for the input image. In addition, not using an explicit interpolation filter means that the network implicitly learns the processing necessary for \ac{SR}. Thus, the network is capable of learning a better and more complex \ac{LR} to \ac{HR} mapping compared to a single fixed filter upscaling at the first layer. This results in additional gains in the reconstruction accuracy of the model as shown in Sec.~\ref{subsection:accuracy} and Sec.~\ref{subsection:video}.
\end{itemize}

We validate the proposed approach using images and videos from publicly available benchmarks datasets and compared our performance against previous works including \cite{dong2015image,chen2015trainable,schulter2015fast}. We show that the proposed model achieves state-of-art performance and is nearly an order of magnitude faster than previously published methods on images and videos.

\section{Method}

The task of \ac{SISR} is to estimate a \ac{HR} image $\networkoutput$ given a \ac{LR} image $\lrimage$ downscaled from the corresponding original \ac{HR} image $\hrimage$. The downsampling operation is deterministic and known: to produce $\lrimage$ from $\hrimage$, we first convolve $\hrimage$ using a Gaussian filter - thus simulating the camera's point spread function - then downsample the image by a factor of $r$. We will refer to $r$ as the upscaling ratio. In general, both $\lrimage$ and $\hrimage$ can have $C$ colour channels, thus they are represented as real-valued tensors of size $H\times W \times C$ and $rH \times rW \times C$, respectively.

To solve the \ac{SISR} problem, the SRCNN proposed in \cite{dong2015image} recovers from an upscaled and interpolated version of $\lrimage$ instead of $\lrimage$. To recover $\networkoutput$, a 3 layer convolutional network is used. In this section we propose a novel network architecture, as illustrated in Fig.~\ref{fig:networkstructure}, to avoid upscaling $\lrimage$ before feeding it into the network. In our architecture, we first apply a $l$ layer convolutional neural network directly to the \ac{LR} image, and then apply a sub-pixel convolution layer that upscales the \ac{LR} feature maps to produce $\networkoutput$.

For a network composed of $L$ layers, the first $L-1$ layers can be described as follows:

\begin{align}
	f^{1}(\lrimage; W_1, b_1) &= \nonlinearity\left( \convolution{W_1}{\lrimage} + b_1 \right),\\
	f^{l}(\lrimage; W_{1:l},b_{1:l}) &= \nonlinearity\left( \convolution{W_{l}}{f^{l-1}\left(\lrimage\right)}+ b_{l} \right),
\end{align}

Where $W_l, b_l, l \in (1, L-1)$ are learnable network weights and biases respectively. $W_l$ is a 2D convolution tensor of size $n_{l-1} \times n_{l} \times k_l \times k_l$, where $n_{l}$ is the number of features at layer $l$, $n_0 = C$, and $k_l$ is the filter size at layer $l$. The biases $b_l$ are vectors of length $n_l$. The nonlinearity function (or activation function) $\nonlinearity$ is applied element-wise and is fixed. The last layer $f^L$ has to convert the \ac{LR} feature maps to a \ac{HR} image $\networkoutput$.

\subsection{Deconvolution layer}

The addition of a deconvolution layer is a popular choice for recovering resolution from max-pooling and other image down-sampling layers. This approach has been successfully used in visualizing layer activations \cite{zeiler2014visualizing} and for generating semantic segmentations using high level features from the network \cite{long2014fully}. It is trivial to show that the bicubic interpolation used in SRCNN is a special case of the deconvolution layer, as suggested already in \cite{long2014fully,dong2015image}. The deconvolution layer proposed in \cite{zeiler2011adaptive} can be seen as multiplication of each input pixel by a filter element-wise with stride $r$, and sums over the resulting output windows also known as backwards convolution \cite{long2014fully}.

\subsection{Efficient sub-pixel convolution layer}
\label{subsec:subpixelconvolutionlayer}

\begin{figure}[htbp]
\begin{center}
\includegraphics[width=\linewidth]{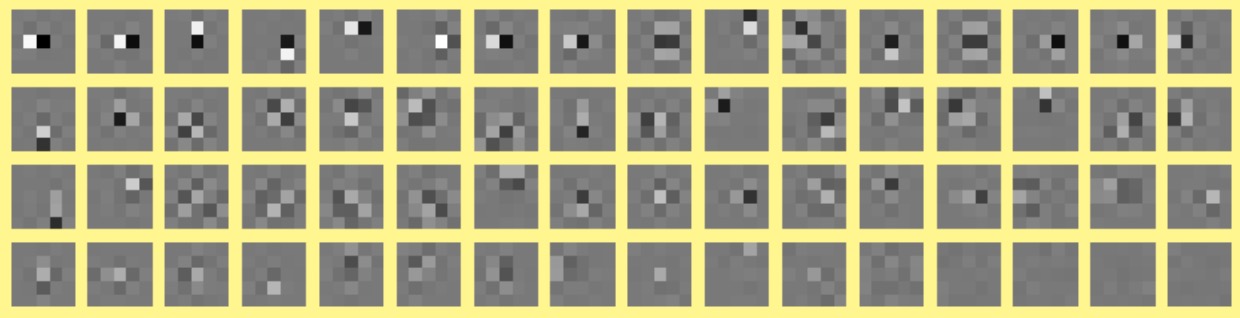}
\caption{The first-layer filters trained on ImageNet with an upscaling factor of 3. The filters are sorted based on their variances.}
\label{fig:firstlayerweights}
\end{center}
\end{figure}

The other way to upscale a \ac{LR} image is convolution with fractional stride of $\frac{1}{r}$ in the \ac{LR} space as mentioned by \cite{long2014fully}, which can be naively implemented by interpolation, {\em perforate} \cite{osendorfer2014image} or un-pooling \cite{zeiler2014visualizing} from \ac{LR} space to \ac{HR} space followed by a convolution with a stride of $1$ in \ac{HR} space. These implementations increase the computational cost by a factor of $r^2$, since convolution happens in \ac{HR} space.

Alternatively, a convolution with stride of $\frac{1}{r}$ in the \ac{LR} space with a filter $W_s$ of size $k_s$ with weight spacing $\frac{1}{r}$ would activate different parts of $W_s$ for the convolution. The weights that fall between the pixels are simply not activated and do not need to be calculated. The number of activation patterns is exactly $r^2$. Each activation pattern, according to its location, has at most $\ceil{\frac{k_s}{r}}^2$ weights activated. These patterns are periodically activated during the convolution of the filter across the image depending on different sub-pixel location: $\Mod{x}{r},\Mod{y}{r}$ where $x, y$ are the output pixel coordinates in \ac{HR} space. In this paper, we propose an effective way to implement the above operation when $\Mod{k_s}{r} = 0$:

\begin{equation}
\networkoutput = f^L(\lrimage) = \periodicshufflingoperator\left ( \convolution{W_L}{f^{L-1}(\lrimage)+b_{L}} \right),
\label{eq:reconstruction}
\end{equation}

where $\periodicshufflingoperator$ is an periodic shuffling operator that rearranges the elements of a $H \times W \times C\cdot r^2$ tensor to a tensor of shape $rH \times rW \times C$. The effects of this operation are illustrated in Fig.~\ref{fig:networkstructure}. Mathematically, this operation can be described in the following way

\begin{equation}
	\periodicshufflingoperator(T)_{x,y,c} = T_{\floor{\nicefrac{x}{r}},\floor{\nicefrac{y}{r}},C \cdot r \cdot mod(y,r) + C \cdot mod(x,r) + c}
	\label{eq:p-shuffling-operator}
\end{equation}

\begin{figure*}[htbp]
\begin{center}
\includegraphics[width=\linewidth]{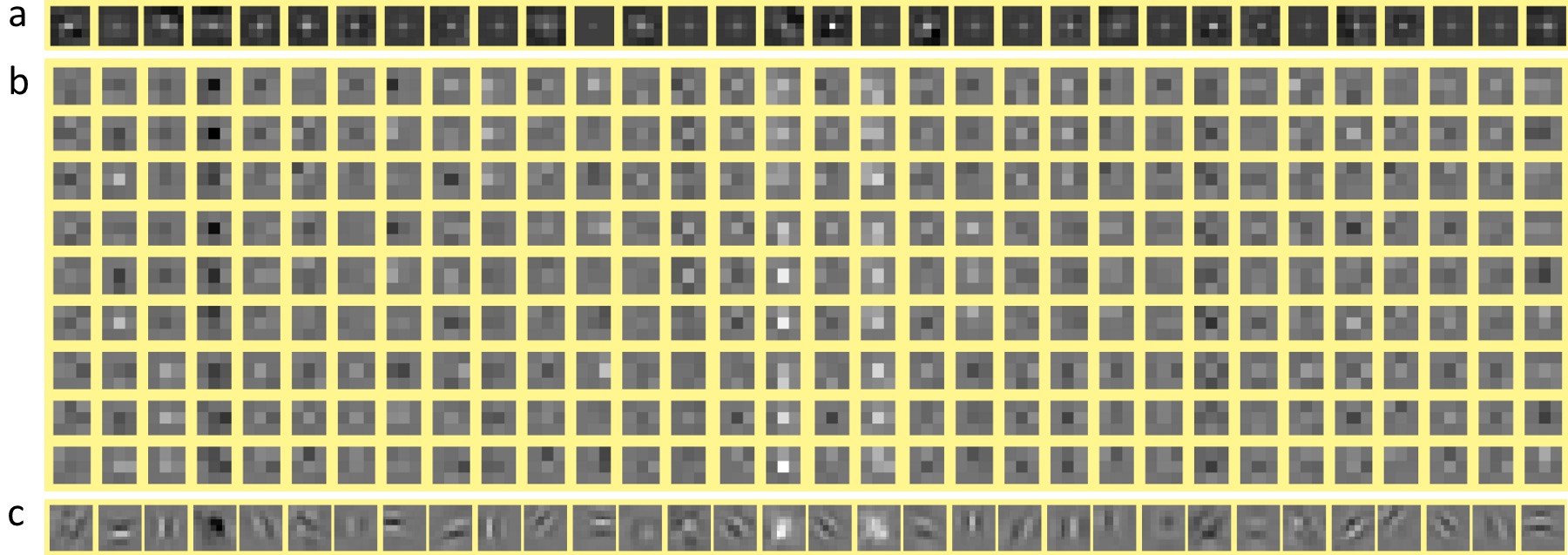}
\caption{The last-layer filters trained on ImageNet with an upscaling factor of 3: (a) shows weights from SRCNN 9-5-5 model \cite{dong2015image}, (b) shows weights from ESPCN (ImageNet $relu$) model and (c) weights from (b) after the $\periodicshufflingoperator$ operation applied to the $r^2$ channels. The filters are in their default ordering.}
\label{fig:lastlayerweights}
\end{center}
\end{figure*}

The convolution operator $W_L$ thus has shape $n_{L-1} \times r^2C \times k_L \times k_L$. Note that we do not apply nonlinearity to the outputs of the convolution at the last layer. It is easy to see that when $k_L = \frac{k_s}{r}$ and $\Mod{k_s}{r} = 0$ it is equivalent to sub-pixel convolution in the \ac{LR} space with the filter $W_s$. We will refer to our new layer as the sub-pixel convolution layer and our network as \ac{ESPCN}. This last layer produces a \ac{HR} image from \ac{LR} feature maps directly with one upscaling filter for each feature map as shown in Fig.~\ref{fig:lastlayerweights}.

Given a training set consisting of \ac{HR} image examples $\hrimage_{n},n=1\ldots N$, we generate the corresponding \ac{LR} images $\lrimage_{n},n=1\ldots N$, and calculate the pixel-wise \ac{MSE} of the reconstruction as an objective function to train the network:

\begin{equation}
	\ell(W_{1:L},b_{1:L}) = \frac{1}{r^2 H W}\sum_{x=1}^{rH} \sum_{x=1}^{rW} \left( \hrimage_{x,y} - f^L_{x,y}(\lrimage)\right)^2
\end{equation}

It is noticeable that the implementation of the above periodic shuffling can be avoided in training time. Instead of shuffling the output as part of the layer, we can pre-shuffle the training data to match the output of the layer before $\periodicshufflingoperator$. Thus our proposed layer is $log_2 r^2$ times faster compared to deconvolution layer in training and $r^2$ times faster compared to implementations using various forms of upscaling before convolution.

\section{Experiments}

\begin{figure*}[htbp]
\begin{center}
\subfloat[Baboon Original]{\includegraphics[width=0.19\textwidth]{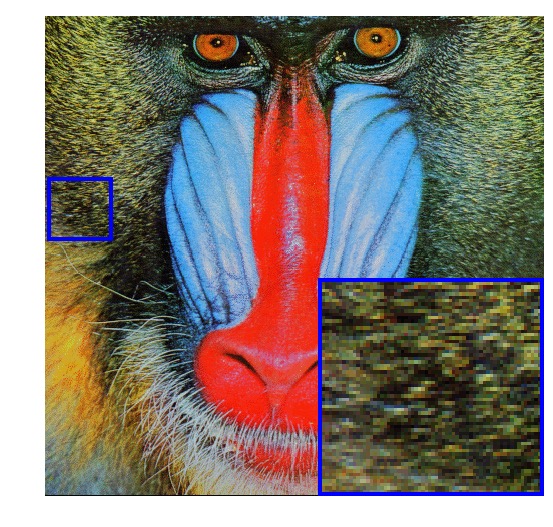}}~
\subfloat[Bicubic / 23.21db]{\includegraphics[width=0.19\textwidth]{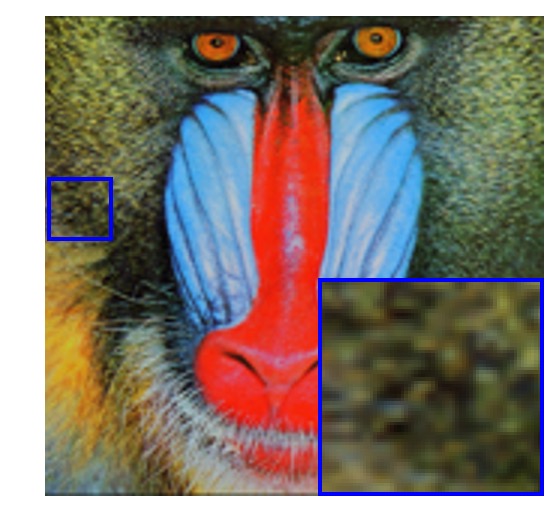}}~
\subfloat[SRCNN \cite{dong2015image} / 23.67db]{\includegraphics[width=0.19\textwidth]{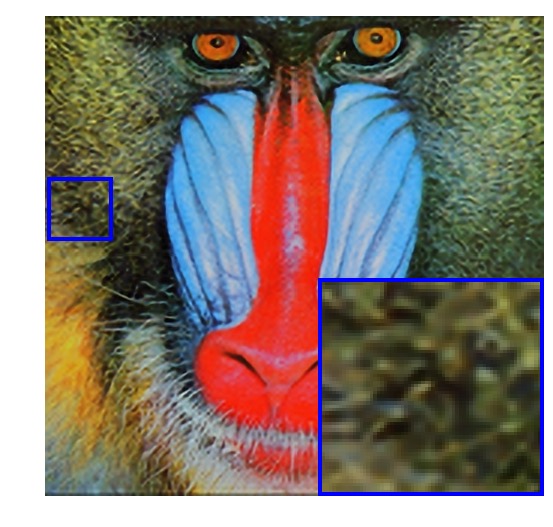}}~
\subfloat[TNRD \cite{chen2015trainable} / 23.62db]{\includegraphics[width=0.19\textwidth]{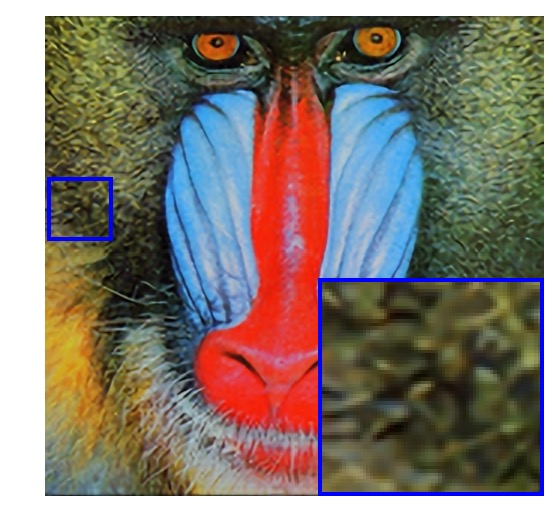}}~
\subfloat[ESPCN / \textbf{23.72db}]{\includegraphics[width=0.19\textwidth]{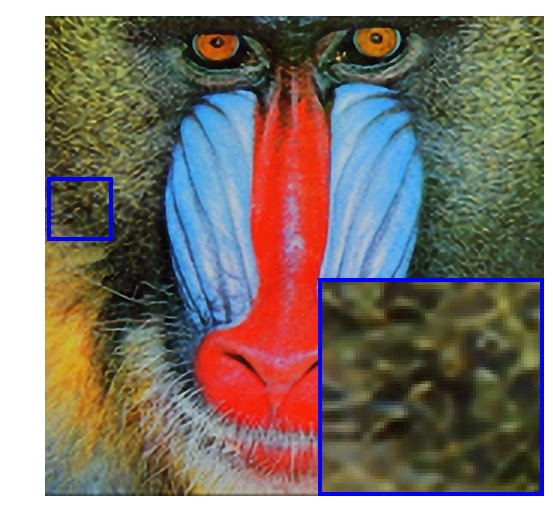}}\\
\subfloat[Comic Original]{\includegraphics[width=0.19\textwidth]{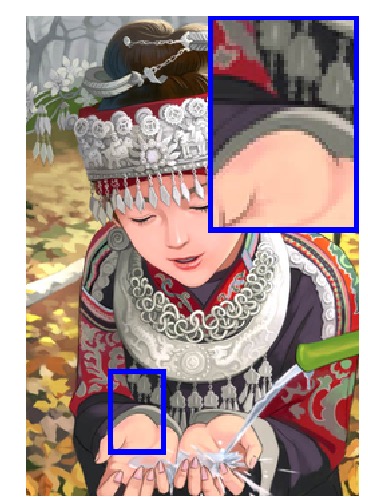}}~
\subfloat[Bicubic / 23.12db]{\includegraphics[width=0.19\textwidth]{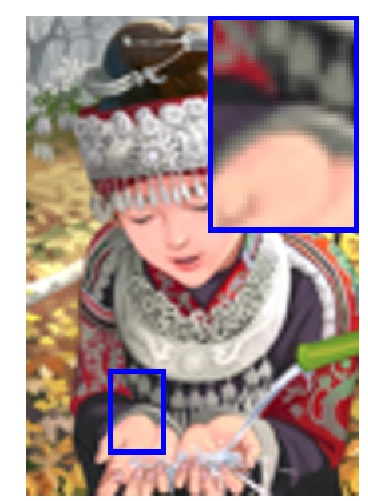}}~
\subfloat[SRCNN \cite{dong2015image} / 24.56db]{\includegraphics[width=0.19\textwidth]{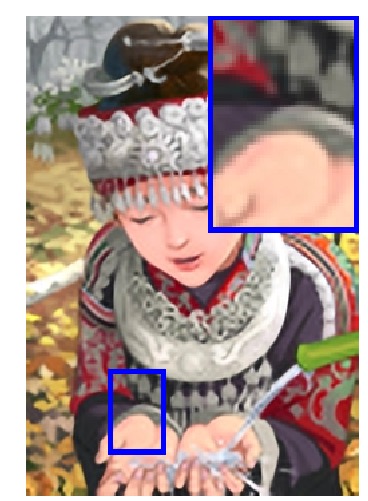}}~
\subfloat[TNRD \cite{chen2015trainable} / 24.68db]{\includegraphics[width=0.19\textwidth]{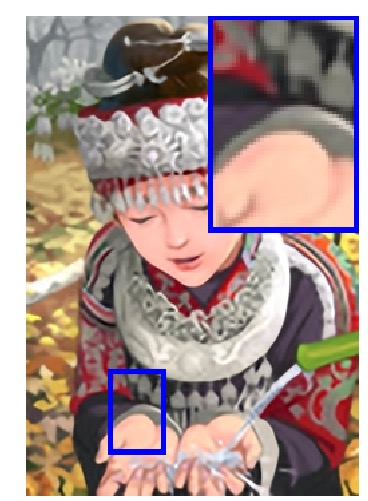}}~
\subfloat[ESPCN / \textbf{24.82db}]{\includegraphics[width=0.19\textwidth]{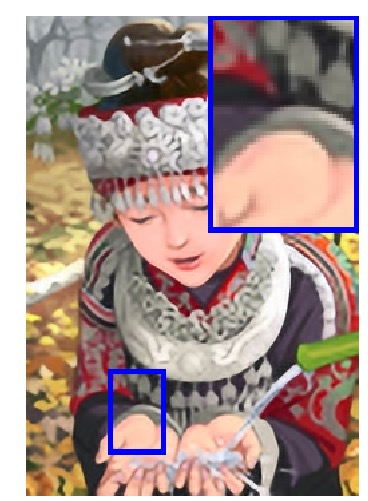}}\\
\subfloat[Monarch Original]{\includegraphics[width=0.19\textwidth]{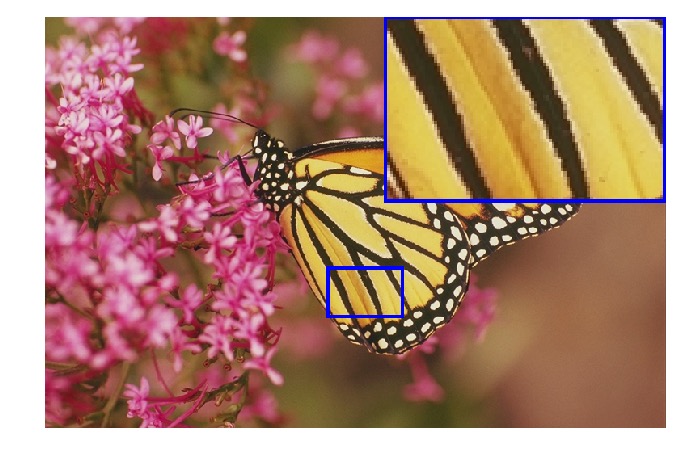}}~
\subfloat[Bicubic / 29.43db]{\includegraphics[width=0.19\textwidth]{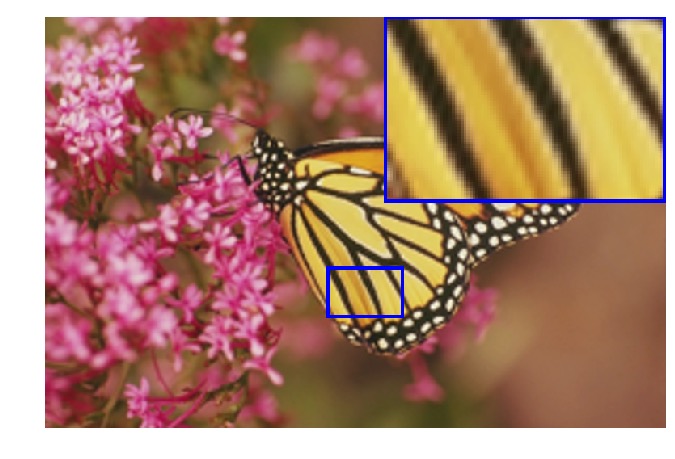}}~
\subfloat[SRCNN \cite{dong2015image} / 32.81db]{\includegraphics[width=0.19\textwidth]{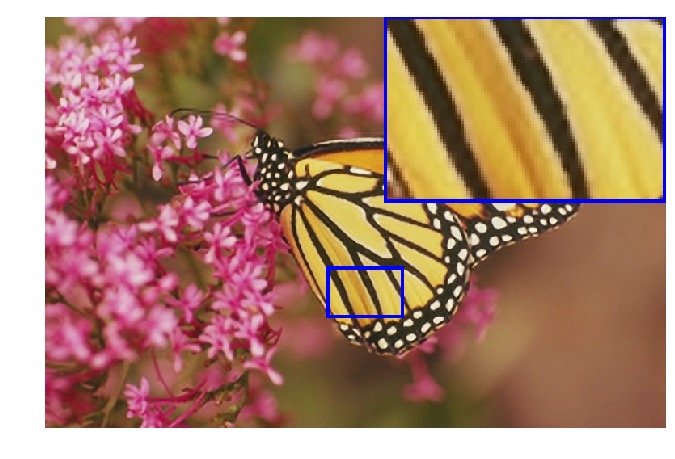}}~
\subfloat[TNRD \cite{chen2015trainable} / 33.62db]{\includegraphics[width=0.19\textwidth]{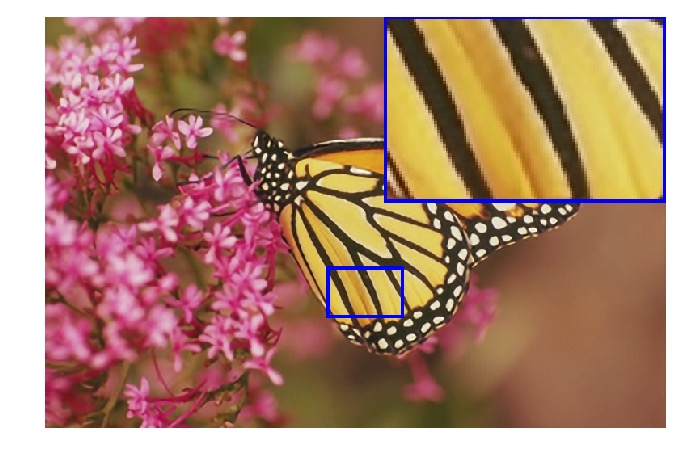}}~
\subfloat[ESPCN / \textbf{33.66db}]{\includegraphics[width=0.19\textwidth]{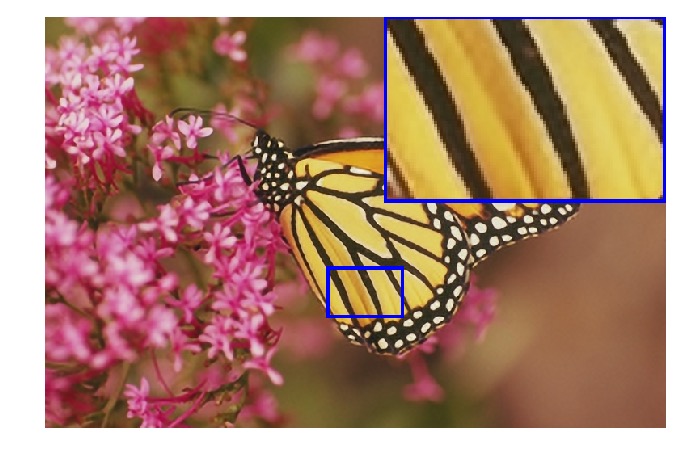}}\\
\caption{Super-resolution examples for "Baboon", "Comic" and "Monarch" from \textbf{Set14} with an upscaling factor of 3. PSNR values are shown under each sub-figure. \label{fig:visualcomparison1}}
\end{center}
\end{figure*}

The detailed report of quantitative evaluation including the original data including images and videos, down-sampled data,  super-resolved data, overall and individual scores and run-times on a K2 GPU are provided in the supplemental material\footnote{Supplemental material \url{https://twitter.box.com/s/47bhw60d066imhh88i2icqnbu7lwiza2}}.

\subsection{Datasets}

During the evaluation, we used publicly available benchmark datasets including the Timofte dataset \cite{timofte2014a+} widely used by \ac{SISR} papers \cite{dong2015image,wang2015deeply,chen2015trainable} which provides source code for multiple methods, 91 training images and two test datasets \textbf{Set5} and \textbf{Set14} which provides 5 and 14 images; The Berkeley segmentation dataset \cite{MartinFTM01} \textbf{BSD300} and \textbf{BSD500} which provides 100 and 200 images for testing and the super texture dataset \cite{dai2015jointly} which provides 136 texture images. For our final models, we use 50,000 randomly selected images from \textbf{ImageNet} \cite{russakovsky2014imagenet} for the training. Following previous works, we only consider the luminance channel in YCbCr colour space in this section because humans are more sensitive to luminance changes \cite{schulter2015fast}. For each upscaling factor, we train a specific network.

For video experiments we use 1080p HD videos from the publicly available \textbf{Xiph} database\footnote{Xiph.org Video Test Media [derf's collection] \url{https://media.xiph.org/video/derf/}}, which has been used to report video \ac{SR} results in previous methods \cite{Takeda2009, liu2011bayesian}. The database contains a collection of $8$ \ac{HD} videos approximately $10$ seconds in length and with width and height $1920 \times 1080$. In addition, we also use the \textbf{Ultra Video Group} database\footnote{Ultra Video Group Test Sequences \url{http://ultravideo.cs.tut.fi/}}, containing $7$ videos of $1920 \times 1080$ in size and $5$ seconds in length.

\subsection{Implementation details}

\begin{figure*}[htbp]
\begin{center}
\subfloat[14092 Original]{\includegraphics[width=0.19\textwidth]{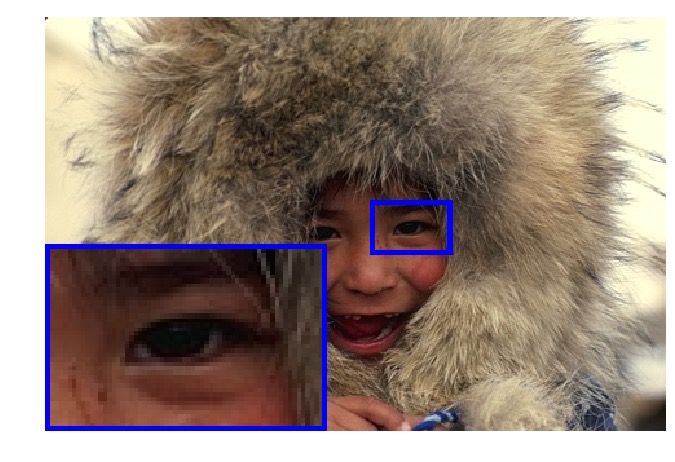}}~
\subfloat[Bicubic / 29.06db]{\includegraphics[width=0.19\textwidth]{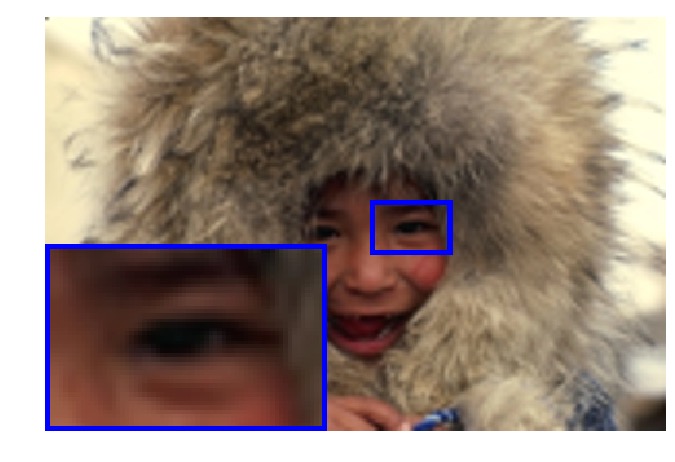}}~
\subfloat[SRCNN \cite{dong2015image} / 29.74db]{\includegraphics[width=0.19\textwidth]{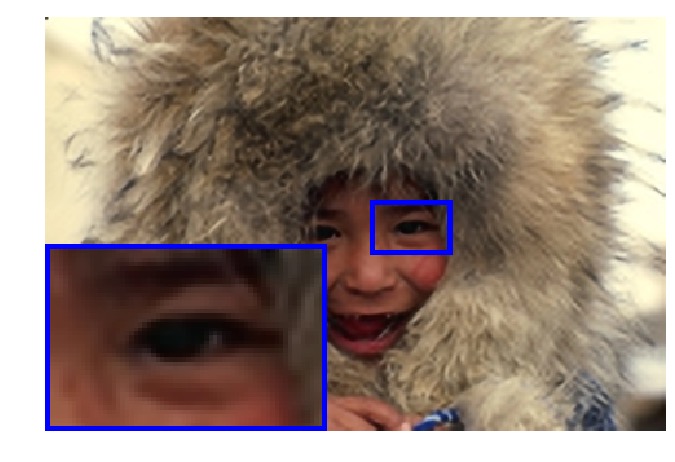}}~
\subfloat[TNRD \cite{chen2015trainable} / 29.74db]{\includegraphics[width=0.19\textwidth]{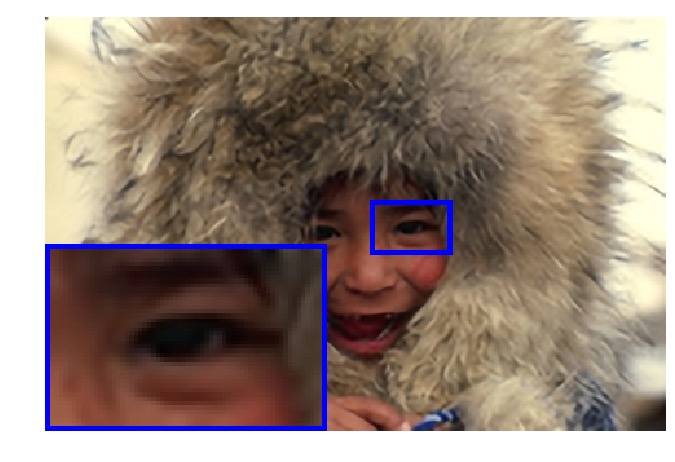}}~
\subfloat[ESPCN / \textbf{29.78db}]{\includegraphics[width=0.19\textwidth]{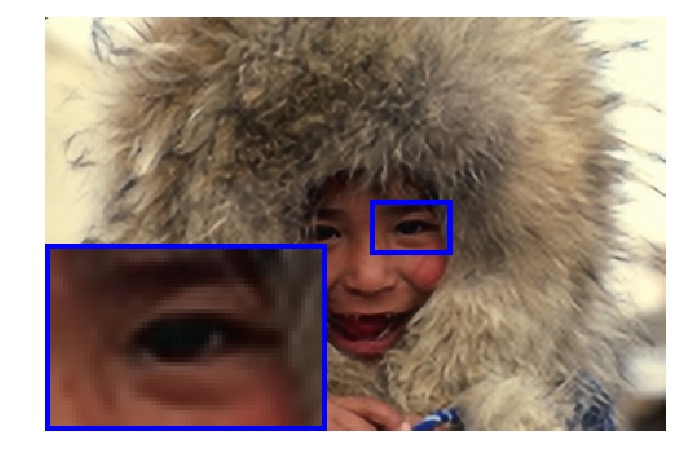}}\\
\subfloat[335094 Original]{\includegraphics[width=0.19\textwidth]{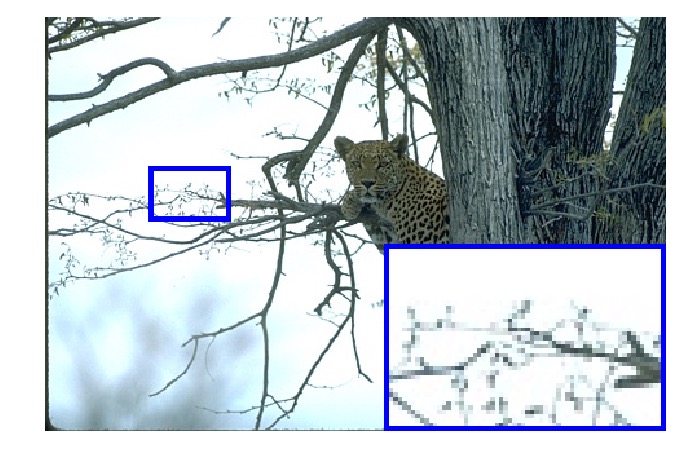}}~
\subfloat[Bicubic / 22.24db]{\includegraphics[width=0.19\textwidth]{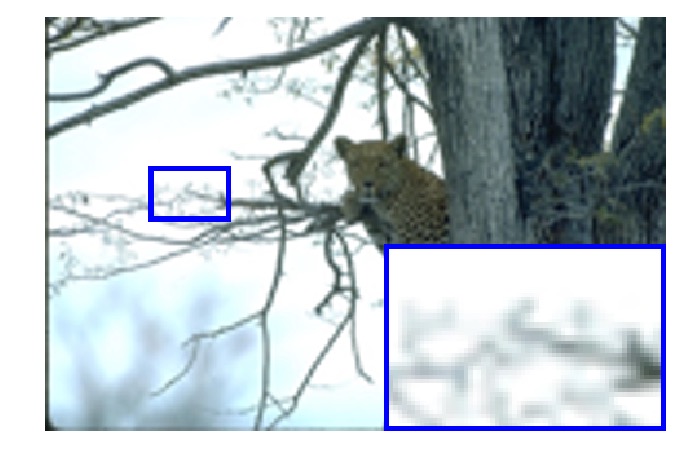}}~
\subfloat[SRCNN \cite{dong2015image} / 23.96db]{\includegraphics[width=0.19\textwidth]{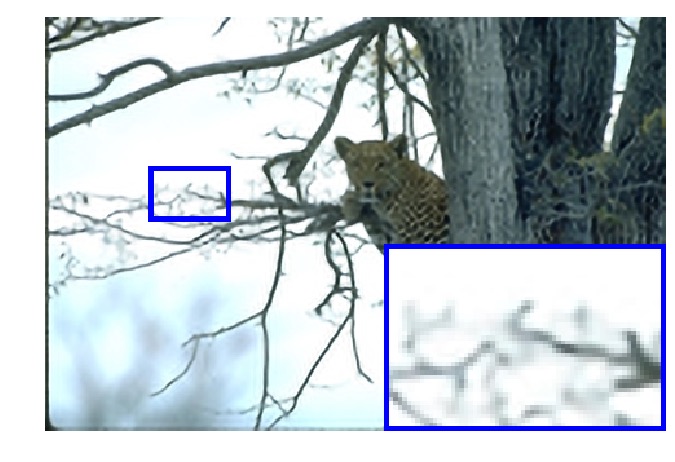}}~
\subfloat[TNRD \cite{chen2015trainable} / \textbf{24.15db}]{\includegraphics[width=0.19\textwidth]{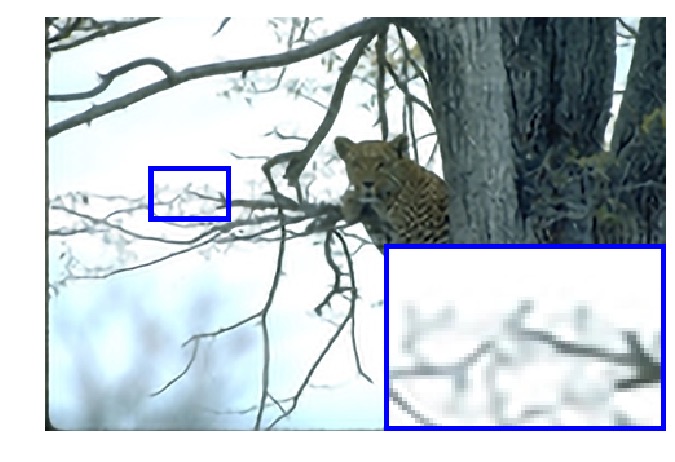}}~
\subfloat[ESPCN / 24.14db]{\includegraphics[width=0.19\textwidth]{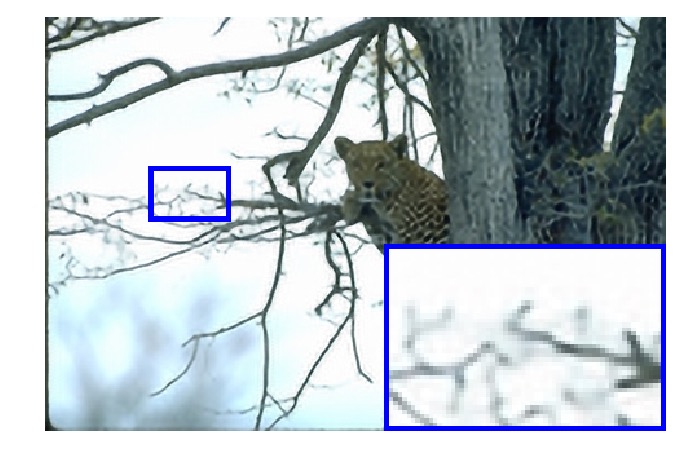}}\\
\subfloat[384022 Original]{\includegraphics[width=0.19\textwidth]{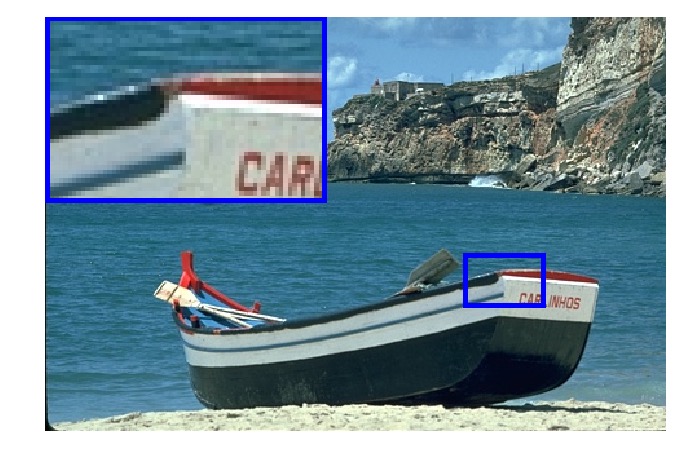}}~
\subfloat[Bicubic / 25.42db]{\includegraphics[width=0.19\textwidth]{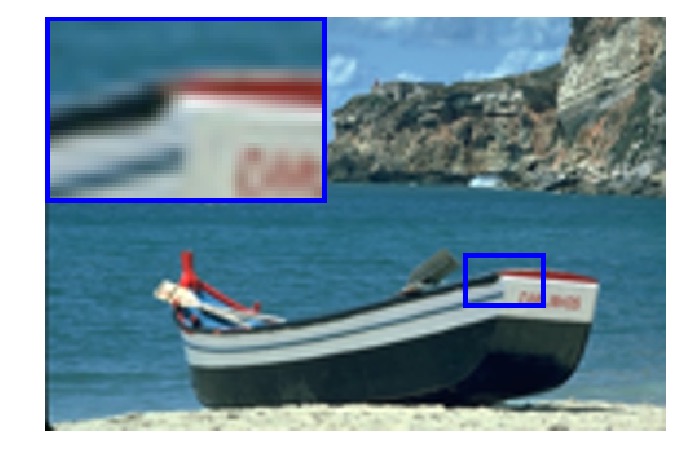}}~
\subfloat[SRCNN \cite{dong2015image} / 26.72db]{\includegraphics[width=0.19\textwidth]{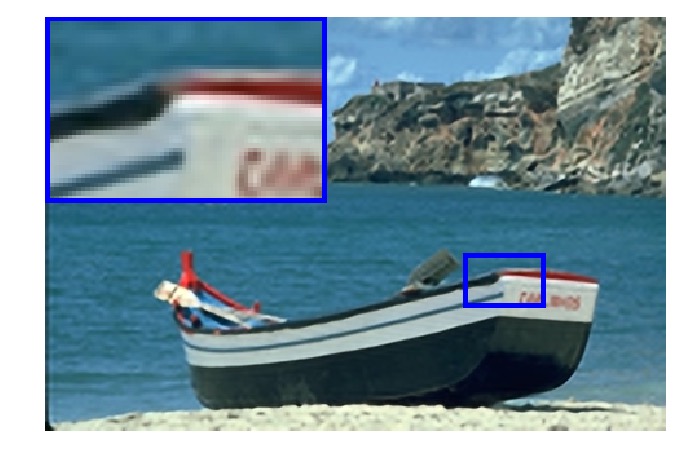}}~
\subfloat[TNRD \cite{chen2015trainable} / 26.74db]{\includegraphics[width=0.19\textwidth]{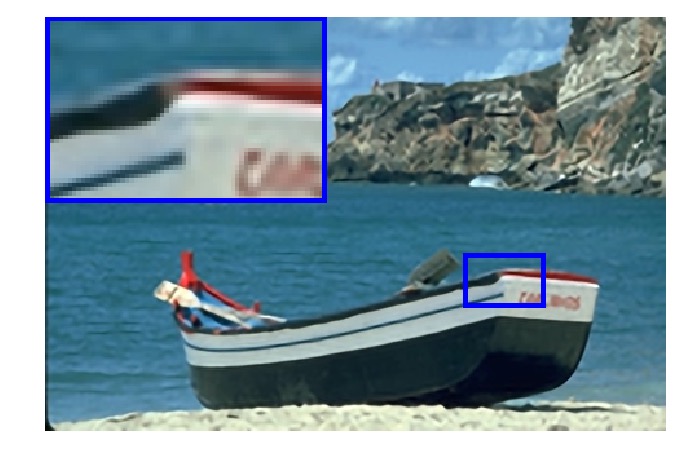}}~
\subfloat[ESPCN / \textbf{26.86db}]{\includegraphics[width=0.19\textwidth]{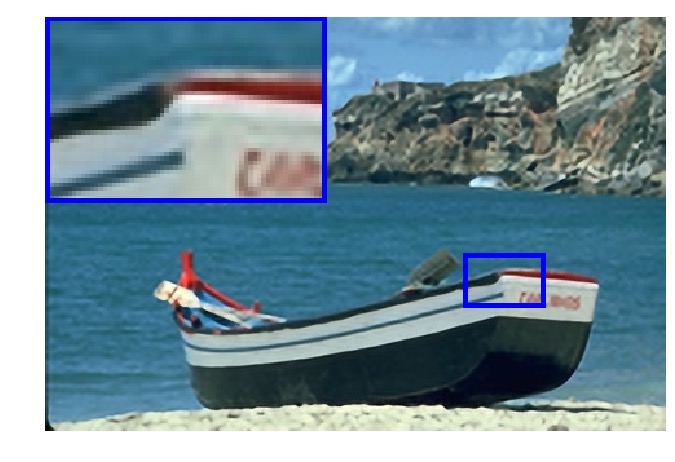}}\\
\caption{Super-resolution examples for "14092", "335094" and "384022" from \textbf{BSD500} with an upscaling factor of 3. PSNR values are shown under each sub-figure. \label{fig:visualcomparison2}}
\end{center}
\end{figure*}

For the \ac{ESPCN}, we set $l = 3$, $(f_1, n_1) = (5, 64)$, $(f_2, n_2) = (3, 32)$ and $f_3 = 3$ in our evaluations. The choice of the parameter is inspired by SRCNN's 3 layer 9-5-5 model and the equations in Sec.~\ref{subsec:subpixelconvolutionlayer}. In the training phase, $17r \times 17r$ pixel sub-images are extracted from the training ground truth images $\mathbf{I}^{HR}$, where $r$ is the upscaling factor. To synthesize the low-resolution samples $\mathbf{I}^{LR}$, we blur $\mathbf{I}^{HR}$ using a Gaussian filter and sub-sample it by the upscaling factor. The sub-images are extracted from original images with a stride of $(17 - \sum{\Mod{f}{2}})\times r$ from $\mathbf{I}^{HR}$ and a stride of $17 - \sum{\Mod{f}{2}}$ from $\mathbf{I}^{LR}$. This ensures that all pixels in the original image appear once and only once as the ground truth of the training data. We choose $tanh$ instead of $relu$ as the activation function for the final model motivated by our experimental results.

The training stops after no improvement of the cost function is observed after 100 epochs. Initial learning rate is set to 0.01 and final learning rate is set to 0.0001 and updated gradually when the improvement of the cost function is smaller than a threshold $\mu$. The final layer learns 10 times slower as in \cite{dong2015image}. The training takes roughly three hours on a K2 GPU on 91 images, and seven days on images from ImageNet \cite{russakovsky2014imagenet} for upscaling factor of 3. We use the \ac{PSNR} as the performance metric to evaluate our models. \ac{PSNR} of SRCNN and Chen's models on our extended benchmark set are calculated based on the Matlab code and models provided by \cite{dong2015image,chen2015trainable}.

\subsection{Image super-resolution results}

\subsubsection{Benefits of the sub-pixel convolution layer}

\begin{table*}
\footnotesize
\begin{center}
\begin{tabular}{|l|c|c|c|c|c|c|}
\hline
Dataset & Scale & SRCNN (91) & ESPCN (91 $relu$) & ESPCN (91) & SRCNN (ImageNet) & ESPCN (ImageNet $relu$) \\
\hline\hline
Set5 & 3 & $32.39$ & $32.39$ & $32.55$ & $32.52$ & $\mathbf{33.00}$ \\
Set14 & 3 & $29.00$ & $28.97$ & $29.08$ & $29.14$ & $\mathbf{29.42}$ \\
BSD300 & 3 & $28.21$ & $28.20$ & $28.26$ & $28.29$ & $\mathbf{28.52}$ \\
BSD500 & 3 & $28.28$ & $28.27$ & $28.34$ & $28.37$ & $\mathbf{28.62}$ \\
SuperTexture & 3 & $26.37$ & $26.38$ & $26.42$ & $26.41$ & $\mathbf{26.69}$ \\
\hline\hline
Average & 3 & $27.76$ & $27.76$ & $27.82$ & $27.83$ & $\mathbf{28.09}$ \\
\hline
\end{tabular}
\end{center}
\caption{The mean PSNR (dB) for different models. Best results for each category are shown in bold. There is significant difference between the \acp{PSNR} of the proposed method and other methods ($p$-value $<$ 0.001 with paired t-test).}
\label{tab:componentsresult}
\end{table*}

In this section, we demonstrate the positive effect of the sub-pixel convolution layer as well as $tanh$ activation function. We first evaluate the power of the sub-pixel convolution layer by comparing against SRCNN's standard 9-1-5 model \cite{dong2014learning}. Here, we follow the approach in \cite{dong2014learning}, using $relu$ as the activation function for our models in this experiment, and training a set of models with 91 images and another set with images from ImageNet. The results are shown in Tab.~~\ref{tab:componentsresult}. \ac{ESPCN} with $relu$ trained on ImageNet images achieved statistically significantly better performance compared to SRCNN models. It is noticeable that \ac{ESPCN} (91) performs very similar to SRCNN (91). Training with more images using \ac{ESPCN} has a far more significant impact on \ac{PSNR} compared to SRCNN with similar number of parameters (+0.33 vs +0.07).

To make a visual comparison between our model with the sub-pixel convolution layer and SRCNN, we visualized weights of our ESPCN (ImageNet) model against SRCNN 9-5-5 ImageNet model from \cite{dong2015image} in Fig.~\ref{fig:firstlayerweights} and Fig.~\ref{fig:lastlayerweights}. The weights of our first and last layer filters have a strong similarity to designed features including the log-Gabor filters \cite{yao2006iris}, wavelets \cite{kingsbury2001complex} and Haar features \cite{viola2001rapid}. It is noticeable that despite each filter is independent in \ac{LR} space, our independent filters is actually smooth in the \ac{HR} space after $\periodicshufflingoperator$. Compared to SRCNN's last layer filters, our final layer filters has complex patterns for different feature maps, it also has much richer and more meaningful representations.

We also evaluated the effect of $tanh$ activation function based on the above model trained on 91 images and ImageNet images. Results in Tab.~~\ref{tab:componentsresult} suggests that $tanh$ function performs better for \ac{SISR} compared to $relu$. The results for ImageNet images with $tanh$ activation is shown in Tab.~~\ref{tab:imageresult}.

\subsubsection{Comparison to the state-of-the-art}
\label{subsection:accuracy}

In this section, we show \ac{ESPCN} trained on ImageNet compared to results from SRCNN \cite{dong2015image} and the \ac{TNRD} \cite{chen2015trainable} which is currently the best performing approach published. For simplicity, we do not show results which are known to be worse than \cite{chen2015trainable}. For the interested reader, the results of other previous methods can be found in \cite{schulter2015fast}. We choose to compare against the best SRCNN 9-5-5 ImageNet model in this section \cite{dong2015image}. And for \cite{chen2015trainable}, results are calculated based on the $7\times7$ $5$ stages model.

Our results shown in Tab.~\ref{tab:imageresult} are significantly better than the SRCNN 9-5-5 ImageNet model, whilst being close to, and in some cases out-performing, the \ac{TNRD} \cite{chen2015trainable}. Although \ac{TNRD} uses a single bicubic interpolation to upscale the input image to \ac{HR} space, it possibly benefits from a trainable nonlinearity function. This trainable nonlinearity function is not exclusive from our network and will be interesting to explore in the future. Visual comparison of the super-resolved images is given in Fig.~\ref{fig:visualcomparison1} and Fig.~\ref{fig:visualcomparison2}, the \ac{CNN} methods create a much sharper and higher contrast images, \ac{ESPCN} provides noticeably improvement over SRCNN.

\begin{table}
\footnotesize
\begin{center}
\begin{tabular}{|l|c|c|c|c|c|}
\hline
Dataset & Scale & Bicubic & SRCNN & TNRD & ESPCN \\
\hline\hline
Set5 & 3 & $30.39$ & $32.75$ & $\mathbf{33.17}$ & $33.13$ \\
Set14 & 3 & $27.54$ & $29.30$ & $29.46$ & $\mathbf{29.49}$ \\
BSD300 & 3 & $27.21$ & $28.41$ & $28.50$ & $\mathbf{28.54}$ \\
BSD500 & 3 & $27.26$ & $28.48$ & $28.59$ & $\mathbf{28.64}$ \\
SuperTexture & 3 & $25.40$ & $26.60$ & $26.66$ & $\mathbf{26.70}$ \\
\hline\hline
Average & 3 & $26.74$ & $27.98$ & $28.07$ & $\mathbf{28.11}$ \\
\hline\hline
Set5 & 4 & $28.42$ & $30.49$ & $30.85$ & $\mathbf{30.90}$ \\
Set14 & 4 & $26.00$ & $27.50$ & $27.68$ & $\mathbf{27.73}$ \\
BSD300 & 4 & $25.96$ & $26.90$ & $27.00$ & $\mathbf{27.06}$ \\
BSD500 & 4 & $25.97$ & $26.92$ & $27.00$ & $\mathbf{27.07}$ \\
SuperTexture & 4 & $23.97$ & $24.93$ & $24.95$ & $\mathbf{25.07}$ \\
\hline\hline
Average & 4 & $25.40$ & $26.38$ & $26.45$ & $\mathbf{26.53}$ \\
\hline
\end{tabular}
\end{center}
\caption{The mean PSNR (dB) of different methods evaluated on our extended benchmark set. Where SRCNN stands for the SRCNN 9-5-5 ImageNet model \cite{dong2015image}, TNRD stands for the Trainable Nonlinear Reaction Diffusion Model from \cite{chen2015trainable} and ESPCN stands for our ImageNet model with $tanh$ activation. Best results for each category are shown in bold. There is significant difference between the PSNRs of the proposed method and SRCNN ($p$-value $<$ 0.01 with paired ttest)}
\label{tab:imageresult}
\end{table}

\subsection{Video super-resolution results}
\label{subsection:video}

\begin{table}
\footnotesize
\begin{center}
\begin{tabular}{|l|c|c|c|c|}
\hline
Dataset & Scale  & Bicubic & SRCNN & ESPCN \\
\hline\hline
SunFlower & 3 & 41.72 & 43.29 & \textbf{43.36} \\
Station2 & 3 & 36.42  & 38.17  & \textbf{38.32}  \\
PedestrianArea & 3 & 37.65 & 39.21  & \textbf{39.27}  \\
SnowMnt & 3 & 26.00  & 27.23  & \textbf{27.20}  \\
Aspen & 3 & 32.75 & \textbf{34.65} & 34.61  \\
OldTownCross & 3 & 31.20  & 32.44  & \textbf{32.53}  \\
DucksTakeOff & 3 & 26.71 & 27.66  & \textbf{27.69}  \\
CrowdRun & 3 & 26.87  & 28.26  & \textbf{28.39}  \\
\hline\hline
Average & 3 & 32.41  & 33.86  & \textbf{33.92} \\
\hline\hline
SunFlower & 4 & 38.99  & 40.57 & \textbf{41.00} \\
Station2 & 4 & 34.13  & 35.72 & \textbf{35.91} \\
PedestrianArea & 4 & 35.49  & 36.94  & \textbf{36.94} \\
SnowMnt & 4 & 24.14  & 24.87  & \textbf{25.13} \\
Aspen & 4 & 30.06  & 31.51  & \textbf{31.83} \\
OldTownCross & 4 & 29.49  & 30.43  & \textbf{30.54} \\
DucksTakeOff & 4 & 24.85  & 25.44  & \textbf{25.64} \\
CrowdRun & 4 & 25.21  & 26.24  & \textbf{26.40} \\
\hline\hline
Average & 4 & 30.30 & 31.47 & \textbf{31.67} \\
\hline
\end{tabular}
\end{center}
\caption{Results on HD videos from \textbf{Xiph} database. Where SRCNN stands for the SRCNN 9-5-5 ImageNet model \cite{dong2015image} and ESPCN stands for our ImageNet model with $tanh$ activation. Best results for each category are shown in bold. There is significant difference between the PSNRs of the proposed method and SRCNN ($p$-value $<$ 0.01 with paired t-test)}
\label{tab:videoresulthd1}
\end{table}

\begin{table}
\footnotesize
\begin{center}
\begin{tabular}{|l|c|c|c|c|}
\hline
Dataset & Scale  & Bicubic & SRCNN & ESPCN \\
\hline\hline
Bosphorus & 3 & 39.38 & 41.07 & \textbf{41.25} \\
ReadySetGo & 3 & 34.64 & 37.33 & \textbf{37.37} \\
Beauty & 3 & 39.77 & 40.46 & \textbf{40.54} \\
YachtRide & 3 & 34.51 & 36.07 & \textbf{36.18} \\
ShakeNDry & 3 & 38.79 & 40.26 & \textbf{40.47} \\
HoneyBee & 3 & 40.97 & 42.66 & \textbf{42.89} \\
Jockey & 3 & 41.86 & 43.62 & \textbf{43.73} \\
\hline\hline
Average & 3 & 38.56 & 40.21 & \textbf{40.35} \\
\hline\hline
Bosphorus & 4 & 36.47 & 37.53 & \textbf{38.06} \\
ReadySetGo & 4 & 31.69 & 33.69 & \textbf{34.22} \\
Beauty & 4 & 38.79 & 39.48 & \textbf{39.60} \\
YachtRide & 4 & 32.16 & 33.17 & \textbf{33.59} \\
ShakeNDry & 4 & 35.68 & 36.68 & \textbf{37.11} \\
HoneyBee & 4 & 38.76 & 40.51 & \textbf{40.87} \\
Jockey & 4 & 39.85 & 41.55 & \textbf{41.92} \\
\hline\hline
Average & 4 & 36.20 & 37.52 & \textbf{37.91} \\
\hline
\end{tabular}
\end{center}
\caption{Results on HD videos from \textbf{Ultra Video Group} database. Where SRCNN stands for the SRCNN 9-5-5 ImageNet model \cite{dong2015image} and ESPCN stands for our ImageNet model with $tanh$ activation. Best results for each category are shown in bold. There is significant difference between the PSNRs of the proposed method and SRCNN ($p$-value $<$ 0.01 with paired t-test)}
\label{tab:videoresulthd2}
\end{table}

In this section, we compare the ESPCN trained models against single frame bicubic interpolation and SRCNN \cite{dong2015image} on two popular video benchmarks. One big advantage of our network is its speed. This makes it an ideal candidate for video \ac{SR} which allows us to super-resolve the videos frame by frame. Our results shown in Tab.~\ref{tab:videoresulthd1} and Tab.~\ref{tab:videoresulthd2} are better than the SRCNN 9-5-5 ImageNet model. The improvement is more significant than the results on the image data, this maybe due to differences between datasets. Similar disparity can be observed in different categories of the image benchmark as \textbf{Set5} vs \textbf{SuperTexture}.

\subsection{Run time evaluations}
\label{subsection:runtime}

In this section, we evaluated our best model's run time on \textbf{Set14}\footnote{It should be noted our results outperform all other algorithms in accuracy on the larger BSD datasets. However, the use of Set14 on a single CPU core is selected here in order to allow a straight-forward comparison with results from previous published results \cite{schulter2015fast,dong2014learning}.} with an upscale factor of 3. We evaluate the run time of other methods \cite{chang2004super,zeyde2012single,timofte2013anchored} from the Matlab codes provided by \cite{timofte2014a+} and \cite{schulter2015fast}. For methods which use convolutions including our own, a python/theano implementation is used to improve the efficiency based on the Matlab codes provided in \cite{dong2015image,chen2015trainable}. The results are presented in Fig.~\ref{fig:accuracyvscomplexity}. Our model runs a magnitude faster than the fastest methods published so far. Compared to SRCNN 9-5-5 ImageNet model, the number of convolution required to super-resolve one image is $r \times r$ times smaller and the number of total parameters of the model is $2.5$ times smaller. The total complexity of the super-resolution operation is thus $2.5 \times r \times r$ times lower. We have achieved a stunning average speed of $4.7ms$ for super-resolving one single image from \textbf{Set14} on a K2 GPU. Utilising the amazing speed of the network, it will be interesting to explore ensemble prediction using independently trained models as discussed in \cite{Szegedy43022} to achieve better \ac{SR} performance in the future.

We also evaluated run time of 1080 \ac{HD} video super-resolution using videos from the \textbf{Xiph} and the \textbf{Ultra Video Group} database. With upscale factor of 3, SRCNN 9-5-5 ImageNet model takes 0.435s per frame whilst our \ac{ESPCN} model takes only 0.038s per frame. With upscale factor of 4, SRCNN 9-5-5 ImageNet model takes 0.434s per frame whilst our \ac{ESPCN} model takes only 0.029s per frame.
\section{Conclusion}
\label{sec:conclusion}

In this paper, we demonstrate that a non-adaptive upscaling at the first layer provides worse results than an adaptive upscaling for \ac{SISR} and requires more computational complexity. To address the problem, we propose to perform the feature extraction stages in the \ac{LR} space instead of \ac{HR} space. To do that we propose a novel sub-pixel convolution layer which is capable of  super-resolving \ac{LR} data into \ac{HR} space with very little additional computational cost compared to a deconvolution layer \cite{zeiler2011adaptive} at training time. Evaluation performed on an extended bench mark data set with upscaling factor of 4 shows that we have a significant speed ($>10\times$) and performance (+0.15dB on Images and +0.39dB on videos) boost compared to the previous \ac{CNN} approach with more parameters \cite{dong2015image} (5-3-3 vs 9-5-5). This makes our model the first CNN model that is capable of \ac{SR} \ac{HD} videos in real time on a single GPU.

\section{Future work}

A reasonable assumption when processing video information is that most of a scene's content is shared by neighbouring video frames. Exceptions to this assumption are scene changes and objects sporadically appearing or disappearing from the scene. This creates additional data-implicit redundancy that can be exploited for video super-resolution as has been shown in \cite{shahar2011space,liu2011bayesian}. Spatio-temporal networks are popular as they fully utilise the temporal information from videos for human action recognition \cite{Ji2013,tran2015spatio}. In the future, we will investigate extending our \ac{ESPCN} network into a spatio-temporal network to super-resolve one frame from multiple neighbouring frames using 3D convolutions.

{\footnotesize
\bibliographystyle{ieee}
\bibliography{bibliograph}
}

\end{document}